\documentclass{wscpaperproc}
\usepackage{latexsym}
\usepackage{graphicx}
\usepackage{mathptmx}

\usepackage{amsmath}
\usepackage{amsfonts}
\usepackage{amssymb}
\usepackage{amsbsy}
\usepackage{amsthm}

\usepackage[pdftex,colorlinks=true,urlcolor=blue,citecolor=black,anchorcolor=black,linkcolor=black]{hyperref}
\usepackage{caption}
\usepackage{subcaption}

\newtheoremstyle{wsc}
{3pt}
{3pt}
{}
{}
{\bf}
{}
{.5em}
{}

\theoremstyle{wsc}

\begin{document}

%
%

\pagestyle{fancyplain}

\thispagestyle{plain}
\firstPageHead{}

\chead{\fancyplain{}{\itshape Morales-Hernández, Van Nieuwenhuyse, Rojas Gonzalez, Jordens, Witters, and Van Doninck}}

\rhead{}
\cfoot{}
\renewcommand{\headrulewidth}{0pt} 

\makeatletter
\let\@internalcite\cite
\def\cite{\def\@citeseppen{-1000}%
    \def\@cite##1##2{(##1\if@tempswa , ##2\fi)}%
    \def\citeauthoryear##1##2##3{##1 ##3}\@internalcite}
\def\citeNP{\def\@citeseppen{-1000}%
    \def\@cite##1##2{##1\if@tempswa , ##2\fi}%
    \def\citeauthoryear##1##2##3{##1 ##3}\@internalcite}
\def\citeN{\def\@citeseppen{-1000}%
    \def\@cite##1##2{##1\if@tempswa, ##2)\else{}\fi}%
    \def\citeauthoryear##1##2##3{##1 (##3)}\@citedata}
\def\citeA{\def\@citeseppen{-1000}%
    \def\@cite##1##2{(##1\if@tempswa , ##2\fi)}%
    \def\citeauthoryear##1##2##3{##1}\@internalcite}
\def\citeANP{\def\@citeseppen{-1000}%
    \def\@cite##1##2{##1\if@tempswa , ##2\fi}%
    \def\citeauthoryear##1##2##3{##1}\@internalcite}
\def\shortcite{\def\@citeseppen{-1000}%
    \def\@cite##1##2{(##1\if@tempswa , ##2\fi)}%
    \def\citeauthoryear##1##2##3{##2 ##3}\@internalcite}
\def\shortciteNP{\def\@citeseppen{-1000}%
    \def\@cite##1##2{##1\if@tempswa , ##2\fi}%
    \def\citeauthoryear##1##2##3{##2 ##3}\@internalcite}
\def\shortciteN{\def\@citeseppen{-1000}%
    \def\@cite##1##2{##1\if@tempswa, ##2\else{}\fi}%
    \def\citeauthoryear##1##2##3{##2 (##3)}\@citedata}
\def\shortciteA{\def\@citeseppen{-1000}%
    \def\@cite##1##2{(##1\if@tempswa , ##2\fi)}%
    \def\citeauthoryear##1##2##3{##2}\@internalcite}
\def\shortciteANP{\def\@citeseppen{-1000}%
    \def\@cite##1##2{##1\if@tempswa , ##2\fi}%
    \def\citeauthoryear##1##2##3{##2}\@internalcite}
\def\citeyear{\def\@citeseppen{-1000}%
    \def\@cite##1##2{(##1\if@tempswa , ##2\fi)}%
    \def\citeauthoryear##1##2##3{##3}\@citedata}
\def\citeyearNP{\def\@citeseppen{-1000}%
    \def\@cite##1##2{##1\if@tempswa , ##2\fi}%
    \def\citeauthoryear##1##2##3{##3}\@citedata}
%
%
%
\def\@citedata{%
    \@ifnextchar [{\@tempswatrue\@citedatax}%
                  {\@tempswafalse\@citedatax[]}%
}

\def\@citedatax[#1]#2{%
\if@filesw\immediate\write\@auxout{\string\citation{#2}}\fi%
  \def\@citea{}\@cite{\@for\@citeb:=#2\do%
    {\@citea\def\@citea{, }\@ifundefined
       {b@\@citeb}{{\bf ?}%
       \@warning{Citation `\@citeb' on page \thepage \space undefined}}%
{\csname b@\@citeb\endcsname}}}{#1}}%

%
\def\@citex[#1]#2{%
\if@filesw\immediate\write\@auxout{\string\citation{#2}}\fi%
  \def\@citea{}\@cite{\@for\@citeb:=#2\do%
    {\@citea\def\@citea{; }\@ifundefined
       {b@\@citeb}{{\bf ?}%
       \@warning{Citation `\@citeb' on page \thepage \space undefined}}%
{\csname b@\@citeb\endcsname}}}{#1}}%

%
\def\@biblabel#1{}
\makeatother



\newdimen\bibindent
\bibindent=0.0em
\def\thebibliography#1{\section*{\refname}\list
   {}{\settowidth\labelwidth{[#1]}
   \leftmargin\parindent
   \itemindent -\parindent
   \listparindent \itemindent
   \itemsep 0pt
   \parsep 0pt}
   \def\newblock{}
   \sloppy
   \sfcode`\.=1000\relax}


\setlength{\baselineskip}{12.7pt}

\title{MULTI-OBJECTIVE SIMULATION OPTIMIZATION OF THE ADHESIVE BONDING PROCESS OF MATERIALS}


\author{Alejandro Morales-Hernández\\
Inneke Van Nieuwenhuyse\\
Sebastian Rojas Gonzalez\\[12pt]
Research Group Logistics, Data Science Institute \\
University of Hasselt\\
Martelarenlaan 42\\B-3500, Hasselt, BELGIUM\\
\and
Jeroen Jordens\\
Maarten Witters\\
Bart Van Doninck\\[12pt]
ProductionS\\Flanders Make\\
Oude Diestersebaan 133\\
3920, Lommel, BELGIUM\\
}

\maketitle

\section*{ABSTRACT}
Automotive companies are increasingly looking for ways to make their products lighter, using novel materials and novel bonding processes to join these materials together. Finding the optimal process parameters for such  adhesive bonding process is challenging. 
In this research, we successfully applied Bayesian optimization using Gaussian Process Regression and Logistic Regression, to efficiently (i.e., requiring few experiments) guide the design of experiments to the Pareto-optimal process parameter settings. 

\section{INTRODUCTION}
The adhesive bonding process we consider consists of different steps: first, plasma is applied to the materials, next glue is applied, and finally the materials are cured in an oven. The optimization problem is bi-objective: the goal is to find the plasma process settings that maximize break strength, while minimizing the associated production costs. We consider six process parameters: 1) whether the materials are pre-processed, 2) the power setting of the plasma torch, 3) the speed at which the plasma torch moves over the sample, 4) the distance between the plasma torch nozzle and the sample, 5) the number of passes of the plasma torch over the sample, and 6) the time between the plasma treatment and the application of the glue.


Real-life experiments measure the break strength at which the components come apart, the type of failure (substrate, cohesive, or adhesive failure) and the occurrence of visual damage. Importantly, process settings that lead to a substrate or cohesive failure or that imply visual damage should be avoided. The break strength is noisy: a given set of process parameters may yield different values in different experiments.
As real-life physical process experiments are time expensive (requiring hours for a single replication), a computer simulator of the adhesive bonding process was developed by the Joining \& Materials Lab of Flanders Make (flandersmake.be/en) to facilitate the development and assess the efficiency of the algorithm. All results presented here are thus obtained with this simulator.

\section{METHOD}
\label{sec:method}


We start with an initial latin hypercube sample consisting of $n_0$ parameter sets, which we evaluate using the simulator. We use augmented Tchebycheff scalarization to transform the two objectives into a single objective, and apply Gaussian Process Regression (GPR or kriging) with heteroscedastic noise \cite{ankenman2008stochastic} to model this  function. Next, the algorithm iteratively selects a single new design (referred to as ``infill point") to evaluate, by maximizing an infill criterion (using the Particle Swarm optimization algorithm). Our infill criterion consists (for any arbitrary point) of the Modified Expected Improvement \cite{gonzalez2020multiobjective} multiplied with the Probability of Feasibility of the point (estimated by a Logistic Regression Classifier, LRC \cite{friedman2001elements}). The simulation results of the infill point are then used to update the GPR and LRC models, and the algorithm continues to iterate until a stopping criterion is reached.



\section{RESULTS AND CONCLUSIONS}
\label{sec:results}

\begin{figure}[!htbp]
\centering
\begin{subfigure}[b]{0.48\textwidth}
\centering
\includegraphics[width=8cm]{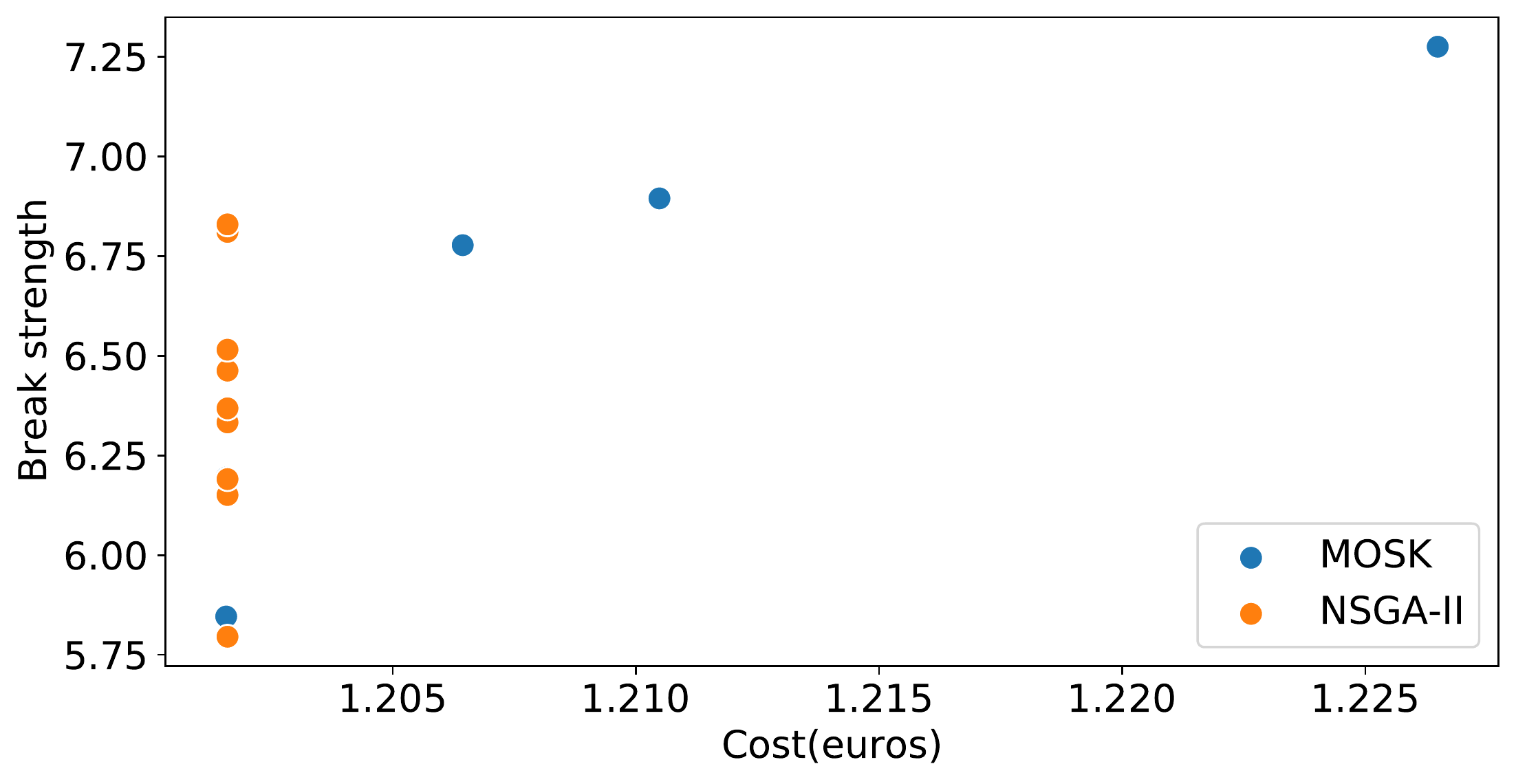} 
\caption{Break strength and cost of the adhesive bonding process: Pareto-optimal solutions.}
\label{fig:PF}
\end{subfigure}
\hfill
\begin{subfigure}[b]{0.48\textwidth}
\centering
\includegraphics[width=8cm]{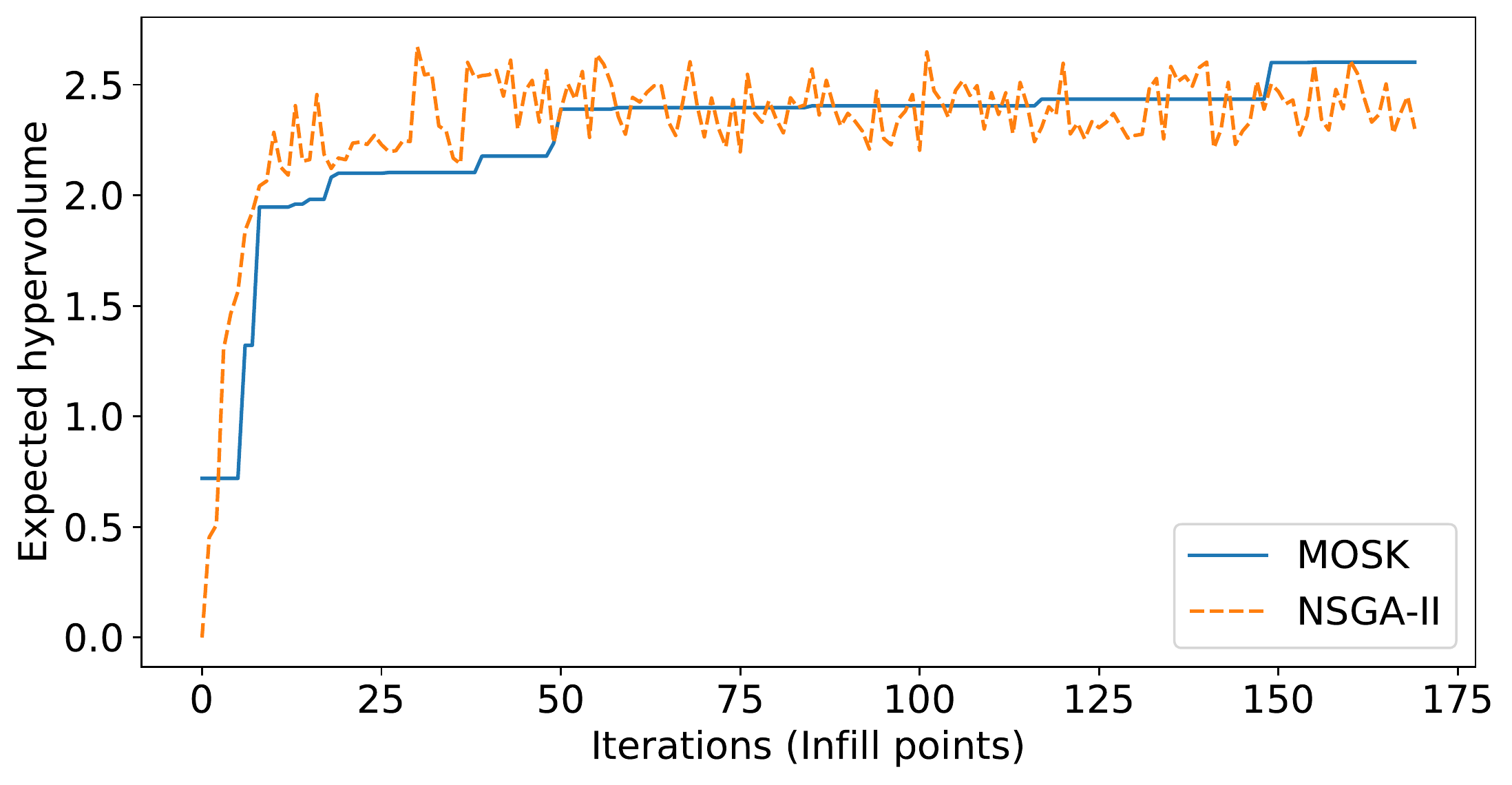} 
\caption{Evolution of the hypervolume of the Pareto-optimal solutions.}
\label{fig:HVf}
\end{subfigure}
\caption{Multi-objective optimization results (only feasible Pareto solutions are considered).}
\label{fig:results_plot}
\end{figure}

We applied a Latin Hypercube sampling of $n_0=30$ initial design points. Figure \ref{fig:PF} shows the Pareto optimal solutions obtained with our algorithm (Multi-objective Stochastic Kriging - MOSK) (after 170 iterations), compared with those  obtained with the Nondominated Sorting Genetic Algorithm II (NSGA-II) (using a population size of 30, and allowing for 170 generations).
As shown in Figure \ref{fig:HVf}, our algorithm converges to a hypervolume that is superior to the one obtained with NSGA-II, requiring only a fraction of the function evaluations (200 vs. 5100). Each generation in NSGA-II is referred to as an \emph{iteration} on the X-axis, and thus entails 30 function evaluations; for MOSK, each iteration requires one function evaluation. 

Our results show that the use of machine learning techniques holds great promise in solving complex and expensive optimization problems, as it allows to obtain high-quality solutions within a smaller number of experiments, compared with a popular and well-known algorithm such as NSGA-II. In this case, the use of the infill criterion allows the algorithm to efficiently search for the Pareto-optimal plasma settings, exploiting the information that has been learned from the already observed process settings through the GPR and LRC models. Future research will focus on the development of an interactive software tool, allowing lab experts to validate the results generated by the algorithm in a real-life test environment, and to apply this type of algorithm also to other process optimization problems.    

\section*{ACKNOWLEDGMENTS}
This research was supported by the Flanders AI Research Program (https://airesearchflanders.be), the Research Foundation-Flanders (FWO Grant 1216021N), and Flanders Make vzw.

\bibliographystyle{wsc}

\bibliography{demobib}

\end{document}